\documentclass[sigconf]{acmart}

\AtBeginDocument{%
  \providecommand\BibTeX{{%
    \normalfont B\kern-0.5em{\scshape i\kern-0.25em b}\kern-0.8em\TeX}}}
\usepackage{multirow}
\usepackage{multicol}

\usepackage{amssymb}
\usepackage{enumitem}
\usepackage{url}
\usepackage{balance} 
\usepackage{xstring}
\usepackage{xspace}
\usepackage{subcaption}
\usepackage{array}
\setlist[itemize,1]{leftmargin=\dimexpr 26pt-2mm}

\copyrightyear{2025} 
\acmYear{2025} 
\setcopyright{acmlicensed}
\acmConference[KDD '25]{Proceedings of the 31st ACM SIGKDD Conference on Knowledge Discovery and Data Mining V.1}{August 3--7, 2025}{Toronto, ON, Canada}
\acmBooktitle{Proceedings of the 31st ACM SIGKDD Conference on Knowledge Discovery and Data Mining V.1 (KDD '25), August 3--7, 2025, Toronto, ON, Canada}
\acmDOI{10.1145/3690624.3709383}
\acmISBN{979-8-4007-1245-6/25/08}

\begin{document}

\newcommand{\model}{LDMapNet-U\xspace}

\title[\model: An End-to-End System for City-Scale Lane-Level Map Updating]{\model: An End-to-End System for City-Scale Lane-Level Map Updating}

\author{Deguo Xia}
\email{xiadeguo@baidu.com}
\authornote{These authors contributed equally to this work.}
\affiliation{%
  \institution{Tsinghua University}
  \state{Beijing}
  \country{China}}
\affiliation{%
  \institution{Baidu}
  \state{Beijing}
  \country{China}}
\orcid{0000-0003-3366-2230}

\author{Weiming Zhang}
\authornotemark[1]
\email{zhangweiming@baidu.com}
\orcid{0000-0003-2609-2807}
\affiliation{%
  \institution{Baidu}
  \state{Beijing}
  \country{China}}

\author{Xiyan Liu}
\authornotemark[1]
\email{liuxiyan@baidu.com}
\orcid{0000-0002-0102-9636}
\affiliation{%
  \institution{Baidu}
  \state{Beijing}
  \country{China}}
  
\author{Wei Zhang}
\authornotemark[1]
\email{zhangwei99@baidu.com}
\orcid{0000-0003-3900-1222}
\affiliation{%
  \institution{Baidu}
  \state{Beijing}
  \country{China}}

\author{Chenting Gong}
\authornotemark[1]
\email{gongchenting@baidu.com}
\orcid{0009-0001-4445-9361}
\affiliation{%
  \institution{Baidu}
  \state{Beijing}
  \country{China}}

\author{Xiao Tan}
\email{tanxiao01@baidu.com}
\orcid{0000-0001-9162-8570}
\affiliation{%
  \institution{Baidu}
  \state{Beijing}
  \country{China}}
  
\author{Jizhou Huang}
\authornote{Corresponding authors.}
\email{huangjizhou01@baidu.com}
\orcid{0000-0003-1022-0309}
\affiliation{%
  \institution{Baidu}
  \state{Beijing}
  \country{China}}

\author{Mengmeng Yang}
\authornotemark[2]
\email{yangmm_qh@tsinghua.edu.cn}
\orcid{0000-0002-3294-6437}
\affiliation{%
  \department{School of Vehicle and Mobility}
  \department{State Key Laboratory of Intelligent Green Vehicle and Mobility}
  \institution{Tsinghua University}
  \state{Beijing}
  \country{China}}
  
\author{Diange Yang}
\email{ydg@mail.tsinghua.edu.cn}
\orcid{0000-0002-0074-2448}
\affiliation{%
  \department{School of Vehicle and Mobility}
  \department{State Key Laboratory of Intelligent Green Vehicle and Mobility}
  \institution{Tsinghua University}
  \state{Beijing}
  \country{China}}

\renewcommand{\shortauthors}{Deguo Xia et al.}

\begin{abstract}
An up-to-date city-scale lane-level map is an indispensable infrastructure and a key enabling technology for ensuring the safety and user experience of autonomous driving systems. In industrial scenarios, reliance on manual annotation for map updates creates a critical bottleneck. Lane-level updates require precise change information and must ensure consistency with adjacent data while adhering to strict standards. Traditional methods utilize a three-stage approach—construction, change detection, and updating—which often necessitates manual verification due to accuracy limitations. This results in labor-intensive processes and hampers timely updates.
To address these challenges, we propose \model, which implements a new end-to-end paradigm for city-scale lane-level map updating. By reconceptualizing the update task as an end-to-end map generation process grounded in historical map data, we introduce a paradigm shift in map updating that simultaneously generates vectorized maps and change information. 
To achieve this, a Prior-Map Encoding (PME) module is introduced to effectively encode historical maps, serving as a critical reference for detecting changes. Additionally, we incorporate a novel Instance Change Prediction (ICP) module that learns to predict associations with historical maps. Consequently, \model~simultaneously achieves vectorized map element generation and change detection. To demonstrate the superiority and effectiveness of \model, extensive experiments are conducted using large-scale real-world datasets. In addition, \model~has been successfully deployed in production at Baidu Maps since April 2024, supporting lane-level map updating for over $360$ cities and significantly shortening the update cycle from quarterly to weekly, thereby enhancing the timeliness and accuracy of lane-level map. The nationwide, high-frequency city-scale lane-level map has been instrumental in the development of the lane-level navigation product serving hundreds of millions of users, while also integrating into the autonomous driving systems of several leading vehicle companies.

\end{abstract}

\ccsdesc[300]{Applied computing~Transportation}
\keywords{Lane-Level Map Updating; End-to-End; Prior Map; Change Detection}

\maketitle

\begin{figure}%
\includegraphics[width=1.0\linewidth]{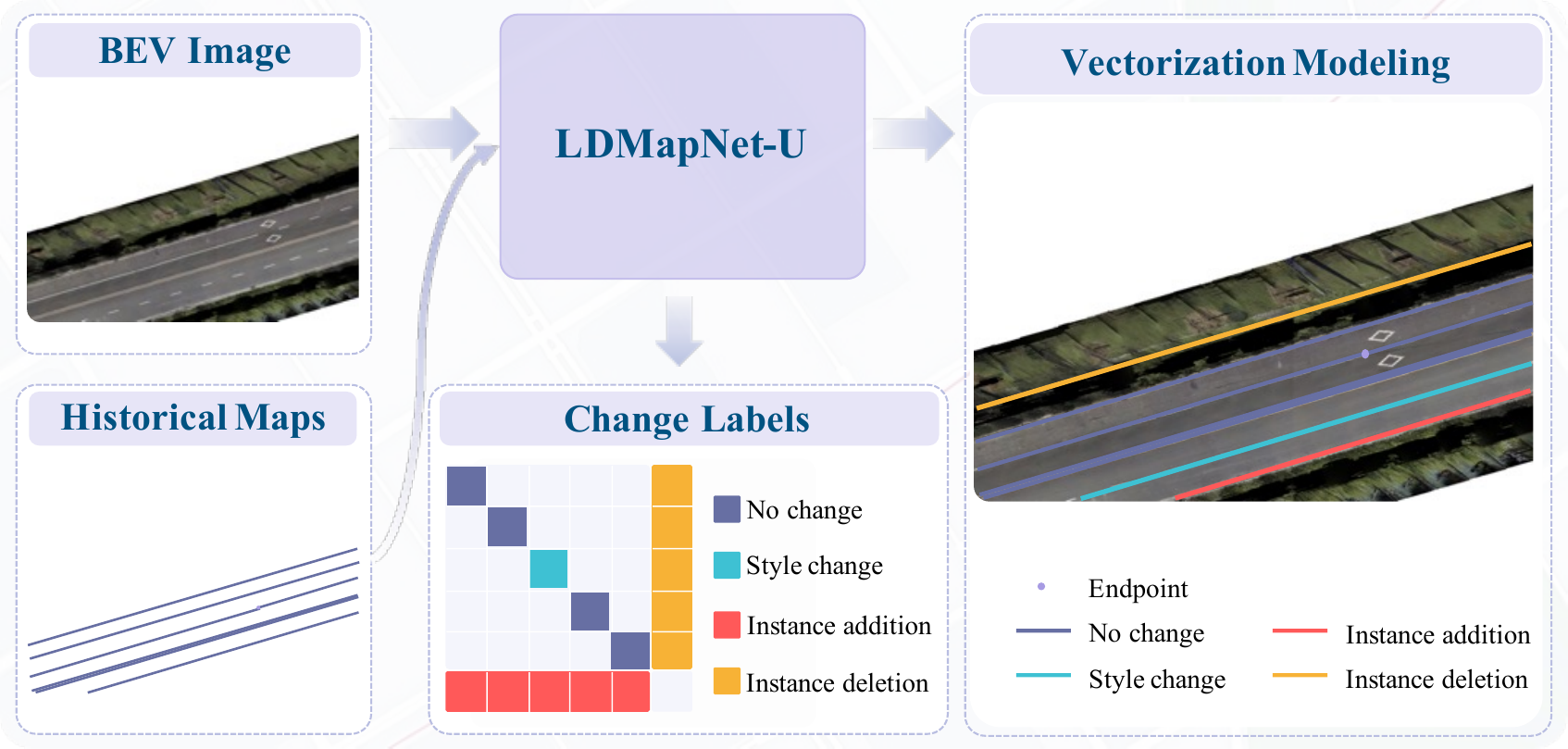}
\vspace{-5mm}
\caption{\model~presents an end-to-end automated industrial-grade approach for lane-level map updating. Our proposed method uses BEV images and historical maps as inputs, achieving end-to-end prediction of vectorized results and change labels through the innovative design of the Prior-Map Encoding (PME) and Instance Change Prediction (ICP) modules. With these advancements, \model~significantly enhances update efficiency and quality.}
\vspace{-4mm}
\label{fig:ad}
\end{figure}

\section{Introduction}

Lane-level map constitutes a foundational component for both navigation and autonomous driving, serving as an indispensable asset in enhancing driving safety and experience by providing comprehensive and highly precise road information. 
However, the widespread deployment of such maps is hindered by two primary challenges: 1) \textit{How to cost-effectively generate lane-level map data for all cities nationwide;} and 2) \textit{How to maintain the currency (a.k.a., freshness, timeliness, or temporal accuracy) of such a massive dataset.} 
To address the initial challenge of generating lane-level map data at scale, we developed DuMapNet~\cite{xia2024dumapnet}, an innovative end-to-end vectorization system for city-scale lane-level map generation. Successfully deployed at Baidu Maps in 2023, DuMapNet achieved a $95\%$ cost reduction while generating detailed lane-level map data covering $361$ cities in China. The dynamic nature of urban environments, characterized by frequent lane configuration changes, poses a significant obstacle to maintaining their accuracy. Effective and scalable solutions for preserving the currency of city-scale lane-level map data remain elusive.

Compared to road-level geo-object or visual change detection~\cite{duarus2022,bu2020mask,varghese2018changenet}, industrial-grade lane-level map updating presents more complex challenges. First, accurately detecting changes at the instance level of individual lanes is fundamental. Second, precise localization of boundaries, including the start and end points of changes, and seamless integration of updates into the existing map are critical. Moreover, the updated map must ensure geometric, topological, and semantic consistency with adjacent unchanged areas. Ultimately, the updated map must adhere to stringent industrial mapping standards, ensuring high accuracy for the rigorous safety and user experience demands of autonomous driving and navigation systems. Traditional methods utilize a three-stage approach—construction, change detection, and updating—which often necessitates manual verification due to accuracy limitations. This labor-intensive process is time-consuming and costly, often involving manual redrawing of road elements according to the strict map-making standards. A more modern approach involves integrating computer vision technology into a multi-stage framework that combines perception and differential analysis components.
While visual models~\cite{tao2020hmsa,liao2022maptr,zhang2023gemap,xia2024dumapnet} are employed to recognize road elements, complex post-processing logic is necessary to compare and identify changes. This multi-step process is prone to error accumulation and lacks automated integration of updated information into the historical maps. The intricate and dynamic urban traffic environment, characterized by frequent vehicle obstructions and lane deterioration, significantly impedes the efficacy of vision-based detection algorithms. Additionally, the accurate and automated incorporation of updated information into lane-level map databases while preserving topological and semantic integrity remains a critical industrial challenge.

To advance the paradigm of lane-level map updating, we present \model, an enhanced version of DuMapNet ~\cite{xia2024dumapnet}, providing an end-to-end automated industrial-grade solution. Unlike traditional multi-stage methods, our model reconceptualizes lane-level map updating as an end-to-end generation process based on historical map data and corresponding latest road observation bird's-eye-view (BEV) images. By taking historical map data and the latest BEV images as input, the model can directly generate standardized vectorized lane-level maps that are consistent with the historical maps, while simultaneously identifying instance changes, thus avoiding the problems of error accumulation and poor generalization in multi-stage methods. To achieve this goal, a novel Prior Map Encoding (PME) module is introduced to effectively encode historical map information as prior knowledge for the model. On one hand, the PME module provides the model with rich historical road prior information, which helps to improve the accuracy of change detection in complex road scenarios; on the other hand, the PME module contributes to improving the geometric, topological, and semantic consistency between the generated map elements and the historical map data. Furthermore, we design an Instance Change Prediction (ICP) module to learn the correspondence and change types between core geographic elements in the latest road observation data and the historical maps. Additionally, by unifying the modeling of polyline-style and polygon-style map elements as a set of points and introducing multi-task joint learning for vectorized map element generation, \model simultaneously achieves end-to-end lane-level vectorized map generation and change detection.

The key contributions to both the research and industrial communities are as follows:

\begin{itemize}
\item \textbf{Potential impact:} We propose an industrial-grade solution, named \model, for city-scale lane-level map updating. \model~has already been deployed in production at Baidu Maps, supporting lane-level map updating for over $360$ cities, accelerating the update cycle from quarterly to weekly. Serving as a foundational component, this city-scale lane-level map empowers navigation for hundreds of millions of users and is integral to the autonomous driving systems of several leading vehicle companies.
\item \textbf{Novelty:} \model~introduces a new paradigm for lane-level map updating that simultaneously generates vectorized maps and change information from BEV images and historical maps. The proposed approach leverage innovative technologies at each stage, including unified vectorization modeling, Prior-Map Encoding (PME) module, and Instance Change Prediction (ICP) module, to achieve a highly automated and cost-effective solution.
\item \textbf{Technical quality:} Extensive qualitative and quantitative experiments on large-scale, real-world datasets demonstrate \model's superiority. Successfully deployed at Baidu Maps, supporting weekly updates for over $360$ cities, underscores \model's practicality and robustness for city-scale lane-level map updating.
\end{itemize}

\section{L\texorpdfstring{D}{D}M\texorpdfstring{\MakeLowercase {ap}}{ap}N\texorpdfstring{\MakeLowercase {et}}{et}-U} \label{section:dumapnet}

\begin{figure*}[ht]
\centering

\begin{subfigure}[b]{0.53\textwidth}
\centering
\includegraphics[width=\textwidth]{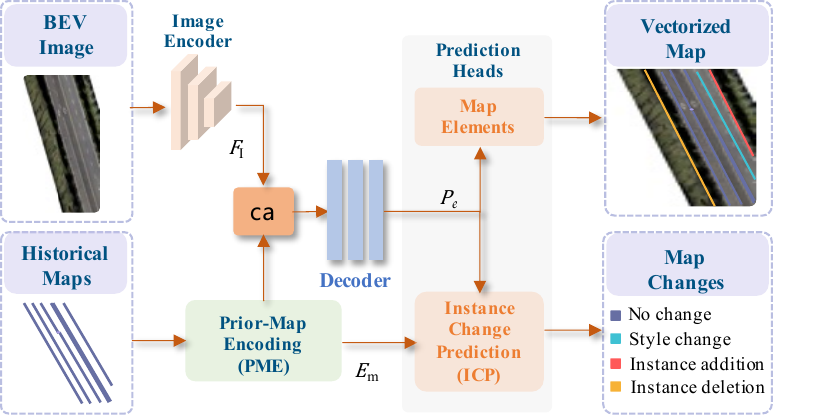}
\caption{}
\label{fig:fig_5a}
\end{subfigure}
\hfill
\begin{subfigure}[b]{0.22\textwidth}
\centering
\includegraphics[width=\textwidth]{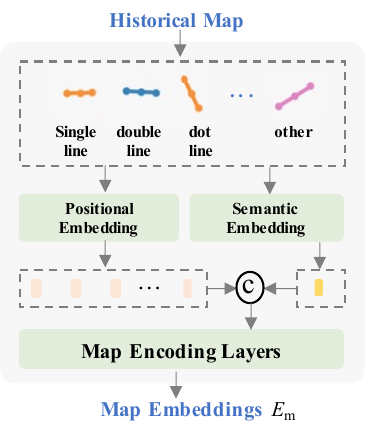}
\caption{}
\label{fig:fig_5b}
\end{subfigure}
\hfill
\begin{subfigure}[b]{0.2\textwidth}
\centering
\includegraphics[width=\textwidth]{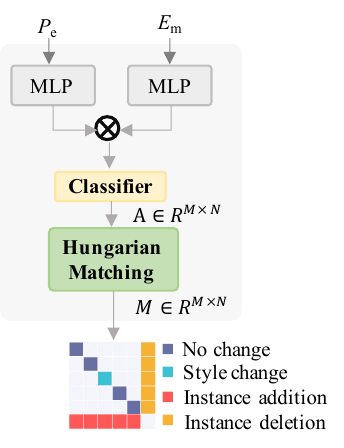}
\caption{}
\label{fig:fig_5c}
\end{subfigure}

\caption{(a) Overall architecture of \model. (b) Our proposed Prior-Map Encoding (PME) module. (c) Our proposed Instance Change Prediction (ICP) module. Please refer to Section \ref{section:dumapnet} for detailed illustrations.}
\label{fig:fig_5}

\end{figure*}

\subsection{Problem Setup}
\label{sec:2.1}
The task of lane-level map updating is defined as follows: given a BEV image $I$ collected from vehicle-mounted sensors and the corresponding historical vectorized map $V_{o}^{I}$ as the input, the network is supposed to generate the updated vectorized map $V^{I}$, while indicating changes to each lane instance, such as style change, instance addition, or deletion. 
Subsequent sections will describe the data preparation and the detailed definition of lane instance changes.

\textbf{Data Preparation. }
As aforementioned, our \model~takes BEV image and vectorized data from historical map database as inputs. To enhance the timeliness of road observation data acquisition, we utilize data from autonomous driving vehicles as a crowdsourced update source. To improve processing efficiency and meet high real-time demands, we have migrated a component of the BEV image creation pipeline from the cloud to the vehicle. This involves initially conducting coarse-grained change detection using on-board perception information from autonomous vehicles, followed by the collection of road images and generation of BEV images for potentially changed road segments in the cloud. Notably, we adopt an adaptive strategy for image acquisition and BEV image generation based on the number of lanes: single-trip collected image data is sufficient for roads with three or fewer lanes, while multi-trips are fused for roads with more than three lanes. 
This approach effectively mitigates the uncertainty inherent in incomplete perception by solely relying on crowdsourced perceived vectorized map elements for map updates.
Note that this part is not the primary focus of this paper and will not be elaborated further. 

Next, we will detail the method of creating the dataset. Firstly, we constructed the dataset of sample pairs, each consisting of three components: the latest BEV image $I$, which provides the most recent observation of the current road scene within region $R$; the historical map $V_o^I$, representing the map information of the region $R$ prior to obtaining BEV image $I$, and the updated map $V^I$, reflecting the map information after incorporating the latest observations from BEV image $I$. Following DuMapNet~\cite{xia2024dumapnet}, the BEV image $I$ with $H \times W$ resolution that covers $H / 25$ meters by $W / 25$ meters of a region with a certain geographic coordinate range. Both $V_o^I$ and  $V^I$ are organized in the form of map meshes, includes all instance information such as geometry, styles, geographic locations, and so on.
Secondly, instances with key geographic features, including geometry (\textit{i.e.}, sets of 2-d points) and style attributes, were extracted from both $V_o^I$ and $V^I$. To align these features with the BEV image, we mapped their geometric shapes to the image's pixel coordinate system. Subsequently, a two-phased labeling process was implemented to identify changes between the historical map $V_o^I$ and updated map $V^I$: an initial automated stage employed instance matching and comparison to generated preliminary change labels, followed by human expert refinement to ensure accuracy. More detailed types of changes are described in the following section. 
Finally, to bolster and validate the model's generalization capabilities, we assembled a comprehensive dataset encompassing diverse urban landscapes across China, characterized by varying road network layouts and complexities.

\textbf{Instance Change Definition.} We define four types of lane instance changes to the map updates as follows:
\begin{itemize}
\item No change: No modifications to existing lane instances.
\item Style change: Alterations to lane instance attributes (\textit{e.g.}, converting solid lane markings to dashed).
\item Instance addition: Inclusion of new lane instances and geometric adjustments (\textit{e.g.}, changes in lane width, curvature, or topological connections).
\item Instance deletion: Removal of lane instances from the historical maps.
\end{itemize}

\subsection{Overall Architecture}
City-scale lane-level map updating presents intricate challenges, demanding precise detection of individual lane instance changes and seamless integration of changes into standardized vectorized maps. To address the limitations of traditional multi-stage methods, our \model proposes an end-to-end framework for city-scale lane-level map updating, realizing practical and effective industrial applications.

Specifically, Figure~\ref{fig:fig_5a} depicts the overall architecture of our proposed \model. 
Taking the latest BEV images and historical maps as input, \model simultaneously predicts vectorized map results and map changes, incorporating innovative Prior-Map Encoding (PME) and Instance Change Prediction (ICP) modules. To extract image features from the input BEV image, we utilize a conventional CNN-based backbone as the image encoder. Subsequent subsections will illustrate each component of the proposed framework.

\subsection{Lane-Level Map Vectorizaiton} 

In our paper, vectorized map $V^I={\lbrace\{P_i, C_i\rbrace\}}_{i=0}^{N}$ is a structured representation of a collection of map elements (\textit{e.g.}, open-shape lane lines), where ${P_i}$ and ${C_i}$ indicate the geometric information and style of each element, and ${N}$ is the number of elements. In order to ensure generality, each element is uniformly defined as a fixed number of vectorized point sets, specifically as $P_i={\lbrace\{p_{ij}\rbrace\}}_{j=0}^{N_p}$, where $N_p$ represents the fixed number of points for each map element. Historical map is a vectorized map at a certain historical moment with same data structurre of $V^I$.  

The primary objective is to convert BEV images into standardized, vectorized map elements in an end-to-end manner. The system begins with an image encoder, which utilizes a conventional CNN-based backbone, such as ResNet50 or HRNet48, to extract features from the input BEV image. The BEV image, collected from vehicle-mounted sensors, provides a comprehensive view of the road network, capturing essential details like lane lines, crosswalks, and other road elements. 

Following the image encoding, the feature decoder adopts the same hierarchical queries architecture as in MapTRv2. Specifically, the hierarchical queries ${Q \in {R^{N \times {N_p}}}}$, includs instance queries and point queries. In addition, the decoder module is composed of several cascaded layers, and each layer integrates a self-attention layer and a cross-attention layer. That design allows hierarchical queries to exchange information across the entire feature space when passing through the self-attention layer, and interact with the BEV features when passing through the cross-attention layer. After obtaining the query embedding ${F_I} \in R^{N \times {N_c}}$ encoded by the decoder, classification and regression networks are respectively used to achieve the style prediction and geometric point set prediction of map elements, where $N_c=256$ represents feature channels. 

\subsection{Prior-Map Encoding (PME)}
To leverage the rich contextual information embedded in historical map data, we introduce the Prior-Map Encoding (PME) module as a novel component of \model. As shown in Figure~\ref{fig:fig_5b}, this module encodes historical map data into a compact representation that captures the geometric and semantic characteristics of road elements and their spatial relationships. 
Specifically, for the $i$-th lane instance, we encode its coordinates $\{h_{ij}\}_{j=1}^{N_p}$ using sine and cosine position encoding to generate the position embedding $\{E^{h}_i\} \in R^{N_c}$, and also encode its category $c_i$ using MLP to generate the semantic embedding $\{E^{h}_i\} \in R^{N_c}$. Naturally, we fuse the positional embeddings and semantic embedding to initially represent historical lane line instances. 
Further we adopt map encoding layers which contains three multi-head self-attention operations to enhance the map representation learning. Taking the advantages of PME, the historical map data is able to serves as an effective prior, providing a reference for the model to effectively detect lane instance changes and update the map accordingly. Subsequently, we adopt a cross-attention operation to fuse map embeddings $E_{m} \in R^{M \times N_c}$ with image embeddings $F_{I} \in R^{N \times {N_c}}$, where $M$ is the number of historical lane instances. Finally, the map-embedding enhanced feature $P_{e} \in R^{N \times {N_c}}$ is fed to the decoder to realize both vectorized map and map change predictions.

\subsection{Instance Change Prediction (ICP)} 

To effectively differentiate lane instance changes, predicting solely based on instance geometry and style is insufficient. Since instance changes inherently relate to historical maps, we reformulate the change prediction task as one of associating predicted instances with existing historical instances. Based on the instance features $P_e$ from the image encoder and the historical instance features $E_m$ from PME, we construct a affinity matrix to represent the associations between the two lane instances, and then effectively generate the corresponding change types. 

\textbf{Associations Matrix Construction. }
As illustrated in Figure~\ref{fig:fig_5c}, we unify the feature space for the predicted and historical instance using distinct MLP networks, followed by an outer product to generate a feature-level association matrix.
Subsequently, a classification network constructs an initial association matrix $\mathcal{A} \in R^{N \times M}$, expressed as follows:
\begin{equation}
\label{equ3.1}
\mathcal{A} = \mathcal{F}(P_{e} \otimes E_{m})
\end{equation}
$\mathcal{F}$ denotes the classification operation.

Finally, during inference, matched pairs of $\mathcal{A}$ with low confidence and inconsistent styles are filtered to enhance the reliability of the association matrix. Meanwhile, based on $\mathcal{A}$, we introduce a Hungarian matching strategy to establish the association matrix $\mathcal{M} \in R^{N \times M}$, ensuring that each predicted lane instance can be matched with at most one historical lane instance and vice versa. Specifically, $M$ and $N$ represent the number of given historical instances and predicted instances, respectively.
An element $\mathcal{M}_{ij}$ equals 1 if the corresponding predicted and historical lane instances are associated, otherwise 0.

\textbf{Instance Change Generation. }
Based on the association matrix $\mathcal{M}$, lane instance changes are categorized as follows:
(1) No change: $\mathcal{M}_{ij}=1$ and the styles of the $i$-th predicted and the $j$-th historical instance are consistent; (2) Style change: $\mathcal{M}_{ij}=1$ and the styles of the $i$-th predicted and the $j$-th historical instance differ;
(3) Instance addition: $\mathcal{M}_{i}=0$ for a predicted instance, indicating that the $i$-th predicted instance has no matching historical instance; (4) Instance deletion: $\mathcal{M}_{j}=0$ for a historical instance, indicating that the $j$-th historical instance does not match any predicted instance.

\begin{table*}
\centering
\caption{Statistics of LD-U dataset.}
\begin{center}
\scalebox{1.0} {
\begin{tabular}{c|cc|cccccc}
\toprule
Dataset &  Mileage (km) & Image & \multicolumn{6}{c}{City} \\
\midrule
Train & \multirow{2}*{9,530} & 142,956 & \multirow{2}*{Guangzhou} & \multirow{2}*{Hangzhou} & \multirow{2}*{Huzhou} & \multirow{2}*{Jining} & \multirow{2}*{Lanzhou} & \multirow{2}*{Tianjin}  \\
Val &  & 15,884 &  &  &  &  \\
\midrule
Test & 360 & 6,000 & Beijing & Chongqing & Dongguan & Harbin & Shaoxing & Yantai \\
\midrule
All & 9,890 & 164,840 & \multicolumn{6}{c}{-} \\
\bottomrule
\end{tabular}
}
\end{center}
\label{table:dataset}
\end{table*}

\subsection{End-to-End Training}

\textbf{Matching.}
Given the task's complexity, which encompasses lane instance coordinate and style prediction alongside change type prediction, we adopt a two-stage matching scheme. The first stage employs hierarchical bipartite matching, similar to MapTR~\cite{liao2022maptr},  to partition samples into positive and negative sets for geometry and style prediction. In the second stage, we leverage known relationships between ground truth and predicted instances, as well as between ground truth and historical instances, to indirectly derive associations between predicted and historical instances. If a predicted lane instance does not match any historical lane instance, it is considered an instance addition and excluded from loss calculation.

\textbf{Map Elements Learning.}
Our approach can be viewed as a multi-task learning framework that encompasses lane instance prediction and lane instance change prediction, as detailed below:
\begin{equation}
\label{equ3}
\mathcal{L} = \alpha \mathcal{L}_{l} + \beta \mathcal{L}_{c}
\end{equation}

Similar to DuMapNet~\cite{xia2024dumapnet}, $\mathcal{L}_{l}$ is composed of three parts: (1) an L1 loss for lane instance coordinate regression, (2) a direction loss based on cosine similarity to enforce lane instance smoothness, (3) an aligned classification loss to ensure accurate instance style and geometric coordinate for each lane instance. For each predicted lane instance, the classification loss is specifically defined as:
\begin{equation} 
\label{equ4}
\mathcal{L}_{aligned\_cls} = \sum_{i=1}^{N_{pos}} (|d_i - p_i|^\gamma) BCE(p_i, d_i) + \sum_{i=1}^{N_{neg}} p_i^\gamma BCE(p_i, 0)
\end{equation}
where $p_i$ is the probability for the $i$-th predicted lane instance, $d_i$ represents the $L_1$ distance between the $i$-th predicted lane instance and its corresponding ground truth, and $N_{pos}$ and $N_{neg}$ denote the number of positive and negative elements, respectively

\textbf{Instance Change Learning.} After filtering out predicted lane instances classified as additions based on the matching strategy, we calculate the classification loss on the association matrix $\mathcal{A}$ formed by the remaining $K$ predicted and $N$ historical lane instance features. Specifically, for each predicted lane instance, it may be associated with any one of the historical lane instances. Therefore, we treat the association relationship of each predicted lane instance as an N-class classification task, expressed as follows:

\begin{equation}
\label{equ5}
\mathcal{L}_{c} = \sum_{i=1}^{K}\sum_{j=1}^{N} \mathcal{L}_{f}(\mathcal{A}_{ij}, \hat{y}_{ij})
\end{equation}
where $\hat{y}_{ij}$ denotes the ground truth label corresponding to $\mathcal{A}_{ij}$, and $\mathcal{L}_{f}$ represents the Focal loss for classification supervision.

\begin{table*}
\caption{To evaluate the performance of map construction, comparisons with state-of-the-art methods on the test set are presented in terms of $R@P_{1,0.8}=80\%$ $(\uparrow)$. $mR$ represents the average of $R@P_{1,0.8}=80\%$.}
\centering
\begin{tabular}{c|cc|cccccc|c|c}
\toprule
\textbf{Method} & \textbf{Backbone} & \textbf{Training Set} & \textbf{Beijing} & \textbf{Harbin} & \textbf{Shaoxing} & \textbf{Yantai} & \textbf{Chongqing} & \textbf{Dongguan} & \textbf{$mR$} & \textbf{FPS} \\
\midrule
MapTR\cite{liao2022maptr} & R50 & LD-U & 68.86 & 69.89 & 73.92 & 66.12 & 68.74 & 66.58 & 69.02 & 29.7 \\
GeMap\cite{zhang2023gemap} & R50 & LD-U & 69.19 & 75.98 & 74.81 & 68.66 & 72.60 & 68.36 & 71.60 & 29.3 \\
DuMapNet\cite{xia2024dumapnet} & R50 & LD-U & 76.93 & 77.43 & 77.04 & 73.23 & 73.94 & 73.18 & 75.29 & 27.9 \\
\midrule
\model~& R50 & LD-U & 84.62 & 85.09 & 84.10 & 85.06 & 82.72 & 82.14 & 83.95 & 27.7 \\
\model~& HR48 & LD-U & 87.68 & 86.58 & 85.26 & 86.98 & 83.86 & 83.58 & 85.65 &  
26.5 \\
\model~& HR48 & LD-U-L & \textbf{93.88} & \textbf{93.14} & \textbf{93.43} & \textbf{94.50} & \textbf{90.07} & \textbf{90.74} & \textbf{92.62} &  26.5 \\
\bottomrule
\end{tabular}
\label{tab:sota_mapconstruction}
\end{table*}

\begin{table*}
\caption{To evaluate the performance of map change detection, comparisons with state-of-the-art methods on the test set are presented in terms of $P_U$ $(\uparrow)$ and $R_U$ $(\uparrow)$. $mP_U$ and $mR_U$ respectively represents the average of $P_U$ and $R_U$. $^{\ast}$ indicates the model is trained on the LD-U-L training set.}
\centering
\begin{tabular}{c|c|>{\centering\arraybackslash}p{2em}>{\centering\arraybackslash}p{2em}|>{\centering\arraybackslash}p{2em}>{\centering\arraybackslash}p{2em}|>{\centering\arraybackslash}p{2em}>{\centering\arraybackslash}p{2em}|p{2em}p{2em}|>{\centering\arraybackslash}p{2em}>{\centering\arraybackslash}p{2em}|>{\centering\arraybackslash}p{2em}>{\centering\arraybackslash}p{2em}|>{\centering\arraybackslash}p{2em}>{\centering\arraybackslash}p{2em}}
\toprule
\multirow{2}*{\textbf{Method}} & \multirow{2}*{\textbf{Backbone}}  & \multicolumn{2}{c|}{\textbf{Beijing}} & \multicolumn{2}{c|}{\textbf{Harbin}} & \multicolumn{2}{c|}{\textbf{Shaoxing}} & \multicolumn{2}{c|}{\textbf{Yantai}} & \multicolumn{2}{c|}{\textbf{Chongqing}} & \multicolumn{2}{c|}{\textbf{Dongguan}} & \multirow{2}*{$mP_U$} & \multirow{2}*{$mR_U$} \\
&& $P_U$ & $R_U$ & $P_U$ & $R_U$ & $P_U$ & $R_U$ & $P_U$ & $R_U$ & $P_U$ & $R_U$ & $P_U$ & $R_U$ &&\\
\midrule
MapTR\cite{liao2022maptr} & R50  & 73.65 & 73.92 & 72.14
 & 69.18 & 69.88 & 67.43 & 73.85 & 72.17 & 72.34 & 68.25 & 72.50 & 68.61 & 72.39 & 69.92  \\
GeMap\cite{zhang2023gemap} & R50  & 75.94 & 74.21 & 75.28 & 73.33 & 71.53 & 68.90 & 75.97 & 74.04 & 74.14 & 70.04 & 73.75 & 70.38 & 74.44 & 71.82 \\
DuMapNet\cite{xia2024dumapnet} & R50  & 78.01 & 76.08 & 76.97 & 75.79 & 72.05 & 70.79 & 78.37 & 76.88 & 75.63 & 72.58 & 77.07 & 75.34 & 76.35 & 74.58 \\
\midrule
\model~& R50 & 79.98 & 85.06 & 77.70 & 83.90 & 76.92 & 83.42 & 79.34 & 84.58 & 76.04 & 82.09 & 77.06 & 83.52 & 77.84 & 83.76 \\
\model~& HR48  & 81.59 & 87.19 & 80.40 & 87.09 & 79.73 & 86.61 & 84.98 & 90.85 & 77.89 & 84.21 & 78.87 & 86.41 & 80.57 & 87.06 \\
\model~$^{\ast}$ & HR48  & \textbf{84.19} & \textbf{91.99} & \textbf{83.46} & \textbf{91.92} & \textbf{82.59} & \textbf{91.46} & \textbf{87.50} & \textbf{94.75} & \textbf{80.56} & \textbf{88.80} & \textbf{81.45} & \textbf{90.24} & \textbf{83.29} & \textbf{91.52} \\
\bottomrule
\end{tabular}
\label{tab:sota_mapchangedetection}
\end{table*}

\section{Experiments}
\subsection{Experimental Settings}

\textbf{Datasets. }
To evaluate the effectiveness of \model, we have collected a large-scale real-world dataset, LD-U, consisting of BEV images, piror map data and ground truth data from twelve cities: Guangzhou, Hangzhou, Huzhou, Jining, Lanzhou, Tianjin, Beijing, Chongqing, Dongguan, Harbin, Shaoxing, and Yantai. These cities exhibit diverse geographical distributions, varying scales and distinct road network configurations. Note that, the first six cities were divided into a training set and a validation set in a $9:1$ ratio, whereas the rest cities were selected as the test set to evaluate the performance of models. Statistically, LD-U contains $164,840$ images, spanning $9,890$ kilometers, with each image at a resolution of $1536 \times 1536$ pixels. More details can be found in Table~\ref{table:dataset}. Additionally, LD-U-L, a dataset containing $1.5$ million samples, has been introduced to demonstrate the effectiveness of \model~on larger-scale data.

\begin{figure*}
\includegraphics[width=0.9\linewidth,trim={0.0cm 0.05cm 0.0cm 0.0cm},clip]{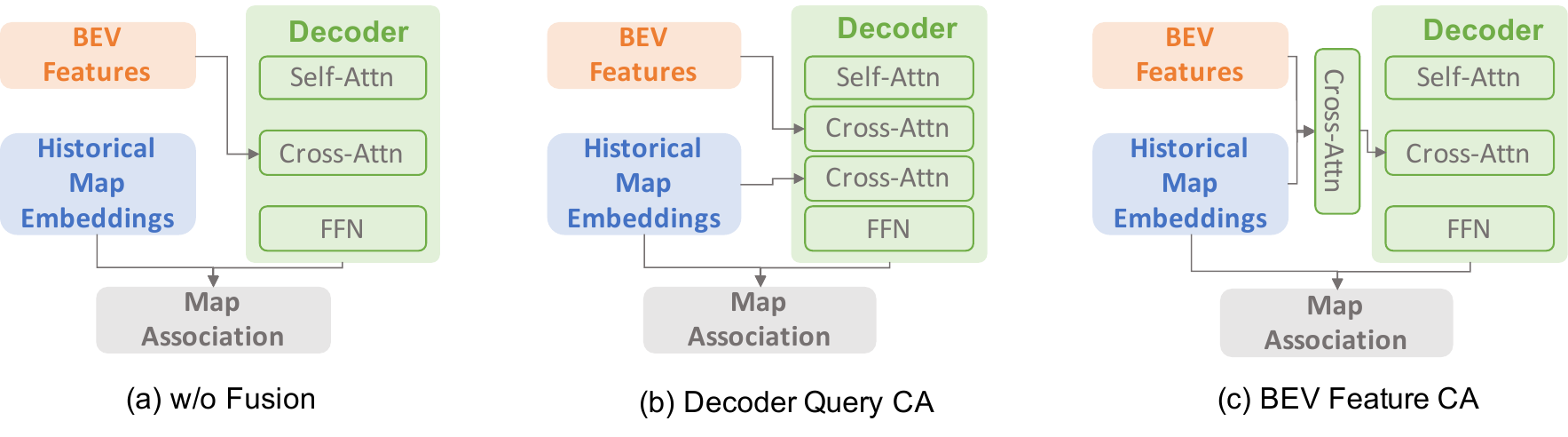}
\caption{Different fusion methods for historical map embeddings. (a) w/o Fusion indicates that the historical map embeddings are not fused with the network before map association prediction. (b) Decoder Query CA refers to the fusion of historical map embeddings with the decoder queries by leveraging multi-head cross-attention. (c) BEV Feature CA refers to the fusion of historical map embeddings with the BEV features by leveraging multi-head cross-attention.}
\label{fig:ablation}
\end{figure*}

\textbf{Evaluation Metrics.}
In the map updating task, both the quality of map construction and the effectiveness of map change detection need to be considered. To evaluate the quality of map construction, we follow DuMapNet \cite{xia2024dumapnet} to use the $R@P_{1,0.8}=80\%$ to represent the recall at $80\%$ precision. For assessing the effectiveness of map change detection, we utilize instance-level change recall ($R_U$) and precision ($P_U$) as the evaluation metrics. In the evaluation process, style change, instance addition, and instance deletion are all considered as map changes.

\textbf{Implementation Details.}
Our model is trained using 16 NVIDIA Tesla V100 GPUs, with a batch size of 16. We utilize the AdamW~\cite{loshchilov2017adamw} optimizer with a weight decay of 0.01, and the initial learning rate is set to $6\times10^{-4}$ with cosine decay. The input images have a resolution of $768\times768$ pixels. For our architecture, we employ ResNet50~\cite{he2016resnet} and HRNet48~\cite{wang2020hrnet} as the backbones. The default number of instance queries, point queries and decoder layers are $50$, $50$ and $6$, respectively. As for hyper-parameters of loss weight, we set $\alpha$ and $\beta$ to $1$ and $1$, respectively. The inference time is measured on a single NVIDIA Tesla V100 GPU with batch size 1.

\subsection{Evaluation}
\textbf{Comparison with Baselines. }
Extensive experiments are conducted on LD-U and LD-U-L. The quantitative comparisons of the quality of the generated elements are summarized in Table~\ref{tab:sota_mapconstruction}. From the results, we observe that \model~consistently brings significant improvements. Taking Beijing as an example, \model~(R50) achieves better performance with $+7.69\% \sim 15.76\%$ recall gains on LD-U, indicating that our model performs better in terms of geometry and style.  In addition, \model~surpasses DuMapNet by a large margin ($+7.06\% \sim 11.83\%$ across six cities). This is reasonable since \model~sufficiently learns from the input historical vectorized map data, facilitating a precise and high-fidelity reconstruction of lane-level elements. Surprisingly, further improvement ($+1.70\%$) is achieved by replacing the backbone with HRNet48, due to the enhanced feature representation. Of particular interest, by augmenting the data volume (\textit{i.e.}, LD-U-L), our method achieves an average recall of $92.62\%$ across six cities, further substantiating the powerful performance and scalability of the model. For computational efficiency, compared to previous methods that only predict maps without detecting lane changes (\textit{e.g.}, MapTR~\cite{liao2022maptr}), the increase in inference cost of our model is nearly negligible.

\begin{table*}
\caption{Comparison of different fusion methods for historical map embeddings. }
\vspace{-3mm}
\centering
\begin{tabular}{c|cc|c|cc}
\toprule
\textbf{Method} & \textbf{Backbone} & \textbf{Training Set} & $mR@P_{1,0.8}=80\%$ & $mP_U$ & $mR_U$ \\
\midrule
w/o Fusion & R50 & LD-U & 75.37 & 77.01 & 76.14 \\
\midrule
Decoder Query CA & R50 & LD-U & 81.63 & 77.53 & 80.81  \\
BEV Feature CA & R50 & LD-U & \textbf{83.95} & \textbf{77.84} & \textbf{83.76}    \\
\bottomrule
\end{tabular}
\label{tab:ablation}
\end{table*}

\begin{table*}
\caption{Ablation study of ICP module.}
\vspace{-3mm}
\centering
\begin{tabular}{c|cc|c|cc}
\toprule
\textbf{Method} & \textbf{Backbone} & \textbf{Training Set} & $mR@P_{1,0.8}=80\%$ & $mP_U$ & $mR_U$ \\
\midrule
w/o ICP & R50 & LD-U & 83.58 & 77.38 & 81.74 \\
\model & R50 & LD-U & \textbf{83.95} & \textbf{77.84} & \textbf{83.76}    \\
\bottomrule
\end{tabular}
\label{tab:ablation_icp}
\end{table*}

Table~\ref{tab:sota_mapchangedetection} presents the comparisons of models in change detection. Notably, since baseline methods, including DuMapNet~\cite{xia2024dumapnet}, cannot predict change labels in an end-to-end manner, we feed their predicted vectorized results into the existing industrial-grade post-processing logic to generate change labels. From the results, we observe that \model~consistently outperforms other methods with validation precision gains of $+1.49\% \sim 5.45\%$ and recall gains of $+9.18\% \sim 13.84\%$. Compared to DuMapNet, \model~ shows the greatest improvement in the instance deletion category, with precision and recall increasing by $1.5\%$ and $13.95\%$, respectively. In the instance addition category, \model~exhibits the smallest improvement, with precision and recall increasing by $1.06\%$ and $7.63\%$, respectively. The main reason for this difference is that for instance addition, it is essential not only to correctly predict the change category but also to ensure that the added instances have the correct geometric information. These results convincingly demonstrates \model's~superior performance in change detection. This can be attributable to the end-to-end paradigm, which directly predicts change labels through the introduced PME and ICP modules, thereby eliminating the reliance on manually set thresholds in post-processing.

\textbf{Analysis of generalization. }
As a deployed algorithm, its generalization capability needed to be thoroughly evaluated. To this end, as shown in Table~\ref{table:dataset}, we select six cities for evaluation that do not overlap with the training set. These cities vary in scale and are distributed across various regions of China. For example, Beijing is a first-tier city, while Shaoxing is a second-tier city; Harbin is located in the northeast of China, whereas Chongqing is in the southwest, reflecting diverse geographical characteristics. From Table~\ref{tab:sota_mapconstruction}, \model~exhibits superior generalization with less fluctuation in performance across six cities. For instance, the maximum deviation of \model~(R50) across the six cities is $2.95\%$, significantly outperforming GeMap's $7.62\%$ and MapTR's $7.80\%$.

\textbf{Ablation Studies. } 
Figure~\ref{fig:ablation} illustrates the three historical map embeddings fusion methods, with their performance shown in Table~\ref{tab:ablation}. According to the experimental results, it is evident that fusing historical map embeddings with the network before map association prediction is crucial, as this enables the incorporation of prior structural and semantic information. Compared to the Decoder Query CA method, our proposed \model~employs the BEV Feature CA method to introduce historical map embeddings at a shallower network level, resulting in improvements of $2.32\%$, $0.31\%$ and $2.9\%$ in $mR@P_{1,0.8}=80\%$, $mP_U$ and $mR_U$, respectively. These findings demonstrate the effectiveness of the fused BEV features, which include not only semantic information from the images but also prior information extracted from the historical maps. In addition, Table~\ref{tab:ablation_icp} illustrates the effectiveness of the proposed ICP module, which pioneers a pattern for end-to-end prediction of change labels, thereby addressing the reliance of earlier methods on complex post-processing logic. As can be seen, ICP has brought improvements in map construction quality ($mR@P_{1,0.8}=80\%$ $+0.37$) and map change detection ($mP_U$ $+0.46$, $mR_U$ $+2.02$).

\begin{figure*}
\includegraphics[width=1.0\linewidth]{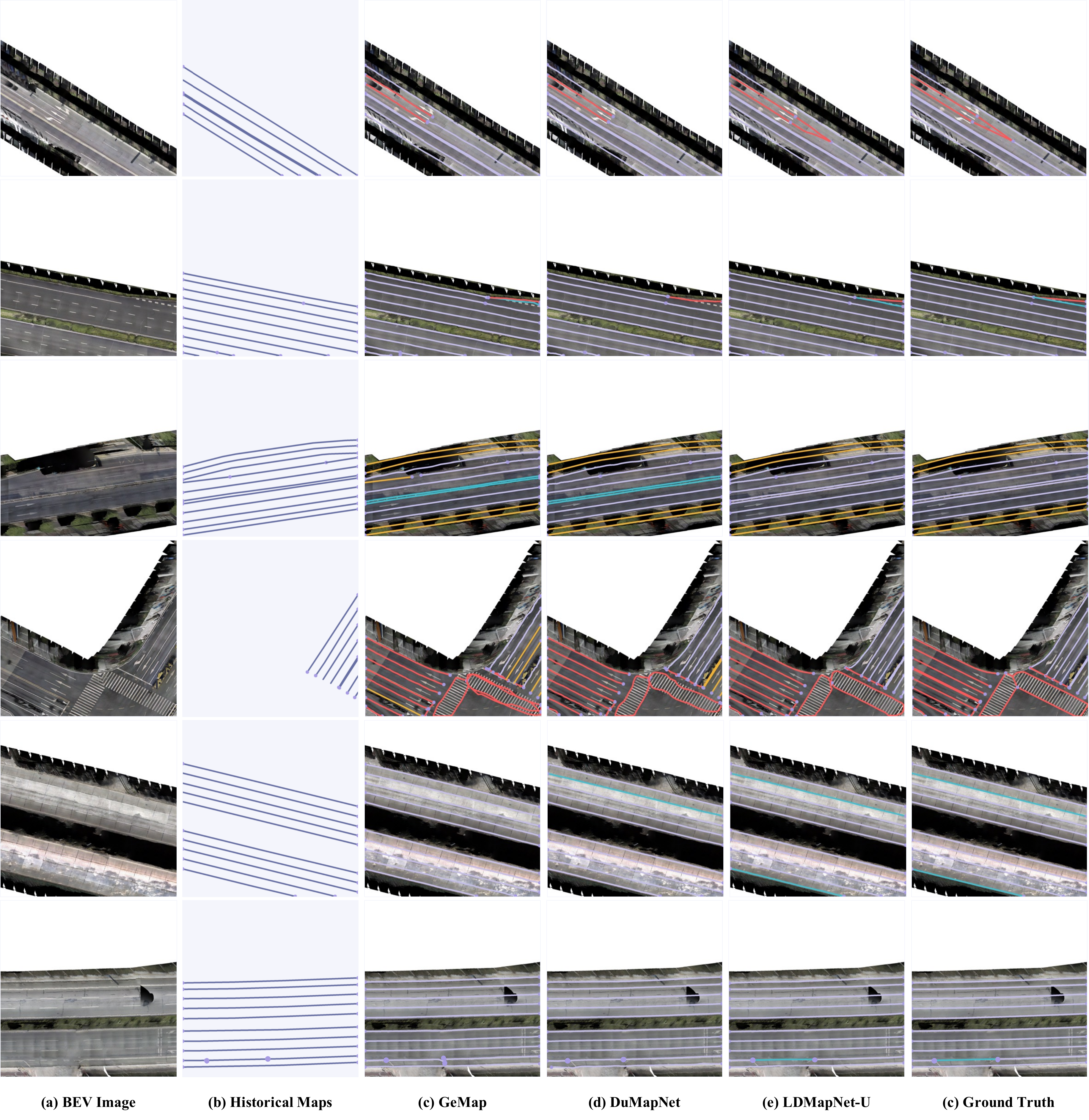}
\caption{Qualitative comparisons of our model with several state-of-the-art models. Instances addition, instance deletion, style change, and no change are highlighted in red, yellow, green, and light purple, respectively. Best viewed in color.}
\label{fig:vis}
\end{figure*}

\subsection{Visualization }
Figure~\ref{fig:vis} illustrates qualitative comparisons between \model~and a range of baseline methods. \model~demonstrates significant advantages in terms of both geometric accuracy and style fidelity of lane instances. This substantial improvement primarily arises from the joint training of the model on BEV images and historical vectorized data, which enables it to adaptively learn the representational relationship between the latent manifolds of multi-source data through the PME module. In essence, BEV images and historical vectorized data exhibit complementary and consistent characteristics in unchanged scenes while demonstrating robust discrimination capabilities in changed portions. As shown in the first row of Figure~\ref{fig:vis}, the upper-right lane changes from two lanes to three lanes. GeMap and DuMapNet fail to accurately output the connection at the location of the lane change, whereas \model successfully predicted the geometry and change labels of the changed lane lines, intuitively demonstrating the model's capability to achieve high-quality end-to-end change label prediction through the PME and ICP modules. Futher, as shown in the third and fifth rows of Figure~\ref{fig:vis}, LDMapNet-U exhibits better performance in occlusion scenarios, potentially due to the introduction of prior map information, which compensates for situations where image features are disrupted.

\section{Related Work}
We provide here some brief context in the fields of map construction and map change detection.

\subsection{Map Construction}
With the development of deep learning and BEV perception \cite{huang2021bevdet}, map construction can be considered as a task of generating maps from sensor observations in a BEV space. The existing methods can be categorized into two types: rasterized map construction and vectorized map construction. Rasterized map construction methods \cite{pan2020vpn, li2022bevformer, hu2021fiery, peng2023bevsegformer, harley2022simple} generate rasterized maps by performing BEV semantic segmentation. To address the limitations of rasterized maps in lacking instance-level structured information, the vectorized map construction methods are proposed and rapidly developed. HDMapNet \cite{li2022hdmapnet} adopts a two-stage approach of segmentation followed by post-processing to generate vectorized instances. As the first end-to-end framework, VectorMapNet \cite{liu2023vectormapnet} utilizes an auto-regressive decoder to predict points sequentially. MapTR \cite{liao2022maptr} proposes a unified shape modeling method based on a parallel end-to-end framework. MapTRv2 \cite{liao2023maptrv2} further introduces auxiliary one-to-many matching and auxiliary dense supervision to speedup convergence. BeMapNet \cite{qiao2023bemapnet} adopts a parameterized paradigm and constructs map elements as piecewise Bezier curves. PivotNet \cite{ding2023pivotnet} utilizes a dynamic number of pivotal points to model map elements, preventing the loss of essential details. GeMap \cite{zhang2023gemap} end-to-end learns Euclidean shapes and relations of map instances beyond basic perception. Unlike existing methods, our proposed \model~not only takes sensor observations as input but also enhances map construction performance by incorporating the corresponding historical map information as guidance.

\subsection{Map Change Detection}
Map change detection aims to detect whether there are any changes in the existing map based on sensor observations of real-world. Early works mostly relied on classical statistical techniques. Pannen et al.~\cite{pannen2019hd} uses a particle filter approach with odometry, Global Navigation Satellite System (GNSS) and landmark readings to obtain distributions evaluated by a number of weak classifiers. Recently, a few efforts have been made to use deep learning for map change detection. Heo et al.~\cite{heo2020hd} directly maps an input image to estimated probabilities of HD map changes based on deep metric learning. Bu et al.~\cite{bu2023toward} projects detections from 2D images onto the BEV space to detect crosswalk changes. Trust but Verify (TbV) \cite{lambert2022trust} is a dataset specifically designed for map change detection, and it has been used to explore deep learning frameworks for map change detection. However, the existing work only focuses on whether the map has changed,  making it difficult to obtain more detailed information such as the location and category of the changes. In order to better support the lane-level map updating task, our proposed \model~can detect lane-level changes, including their locations and categories.

\section{Discussion}
As described in Section~\ref{sec:2.1}, our method employs BEV images as the latest road observation data, rather than other data sources such as vectorized results transmitted from the vehicle or satellite images. This choice is due to the fact that, unlike conventional onboard perception systems, our deployed BEV image creation process can leverage comprehensive regional global information, such as geometric smoothness constraints and semantic associations. Moreover, compared to satellite images, BEV images primarily cover road areas, while possessing the advantages of high flexibility and high resolution. For city-scale lane-level map updating task, BEV images offer advantages such as high timeliness, global consistency, and the ability to overcome inherent challenges of onboard perception, including precision bias, perceptual incompleteness, and dynamic occlusions \cite{chen2024mapcvv,kim2021hd, han2023collaborative}. LDMapNet-U has greatly increased automation in large-scale commercial map systems, cutting operational costs by 95$\%$, reducing the map update cycle from quarterly to weekly. 
Furthermore, high-quality lane-level maps will potentially benefit a range of geospatial-related tasks, such as traffic condition prediction~\cite{dutraffic2022}, estimated time of arrival prediction (ETA prediction, a.k.a., travel time estimation)~\cite{dueta2022, fang2020constgat, fang2021ssml}, road extraction \cite{duare2022}, and geospatial foundation models \cite{huang2022ernie,xiao2024refound}.

Despite the impressive achievements of \model, several challenges worth exploring in the future. For instance, due to the inherent challenges associated with lane-level map, including the demand for high precision and the multitude of map elements, our method relies on high-precision crowdsourced data, such as images collected by autonomous driving vehicles with decimeter-level or higher localization accuracy. The challenge of utilizing lower precision crowdsourced data to meet map-making standards remains an open question. The fusion of multi-source data is expected to be a valuable research area. For instance, combining trajectory data with BEV images could enhance recognition performance in cases where image features are less pronounced. Furthermore, \model~lacks efficiency in updating dynamic events, such as short-term construction or traffic accidents, which often require detection within even hours. To address this challenge, higher-timeliness and lower-cost crowdsourced images or trajectory data may be an effective solution. 

\section{Conclusions}
In this paper, we propose an effective industrial-grade solution for lane-level map updating. Specifically, the proposed \model~employs the Prior-Map Encoding (PME) module to incorporate historical map data as one of its inputs. This approach provides critical reference information for detecting changes while improving the quality of the predicted instances. Additionally, the introduction of the Instance Change Prediction (ICP) module enables the model to directly predict change labels in a learning-based manner, thus eliminating the need for cumbersome post-processing logic. Extensive quantitative and qualitative experiments conducted on the collected large-scale real-world dataset from Baidu Maps demonstrate the effectiveness and superiority of \model. Since its deployment at Baidu Maps in April 2024, \model~has successfully supported weekly updates for over $360$ cities in China, realizing a nationwide, high-frequency lane-level map essential for navigation and autonomous driving. This further demonstrates its cost-effectiveness and practicality as an industrial-grade solution.

\section{Acknowledgments}
This work was supported in part by the National Natural Science Foundation of China (52472449, U22A20104), Beijing Natural Science Foundation (23L10038,L231008), and Beijing Municipal Science and Technology Commission (Z241100003524013, Z241100003524009).

\clearpage

\balance
\bibliographystyle{ACM-Reference-Format}
\bibliography{ref}

\end{document}